\documentclass{llncs}
\usepackage{llncsdoc}
\usepackage{times}
\usepackage{url}
\usepackage{mdwlist}
\usepackage{latexsym}
\usepackage{graphicx}
\usepackage{amssymb}
\usepackage{color}

\usepackage[belowskip=-15pt,aboveskip=2pt]{caption}
\begin{document}

\title{How Easy is it to Learn a Controlled Natural Language for Building a Knowledge Base?}
\author{Sandra Williams, Richard Power and Allan Third}
\institute{The Open University, Walton Hall, Milton Keynes, MK7 6AA,  U.K. \\ 
                \email{sandra.williams@open.ac.uk}}
\maketitle

\begin{abstract}
Recent developments in controlled natural language editors for knowledge engineering (KE) have given rise to expectations that they will make KE tasks more accessible and perhaps even enable non-engineers to build knowledge bases. This exploratory research focussed on novices and experts in knowledge engineering during their attempts to learn a controlled natural language (CNL) known as OWL Simplified English and use it to build a small knowledge base. Participants' behaviours during the task were observed through eye-tracking and screen recordings.

This was an attempt at a more ambitious user study than in previous research because we used a naturally occurring text as the source of domain knowledge, and left them without guidance on which information to select, or how to encode it.
We have identified a number of skills (competencies) required for this difficult task and key problems that authors face.
\end{abstract}

\section{Introduction}

Controlled Natural Language (CNL) has been proposed as a convenient and accessible medium for building knowledge bases such as semantic web ontologies, e.g., ACE \cite{Fuchs2006}, Sidney OWL syntax \cite{Cregan2007}, OSE \cite{Power2012}, CLOnE \cite{Funk2007}, Rabbit \cite{dolbear07} or software requirements specifications \cite{zapata2012}.  CNLs for these tasks are designed to be unambiguously interpreted, usually by machine, into formal languages; consequently, they have been proposed as an alternative to formal representation languages such as the Web Ontology Language (OWL).\footnote{www.w3.org/TR/owl-features/} It has been assumed that since a CNL closely resembles a natural language (NL) it will be easy to learn, especially if the editor has a predictive interface \cite{Schwitter2010}, and thus the task of constructing a knowledge base will be reduced to the task of constructing syntactically correct and semantically plausible CNL sentences.  CNLs have been proposed as particularly useful for non-experts in knowledge representation (KR) languages, enabling them to encode their own domain knowledge into a formal representation, perhaps without any help from a knowledge engineer. 

However, these underlying assumptions have undergone little previous evaluation (see section \ref{relWork}). Thus, the study described here investigated: (i) Is a CNL easy to learn? and (ii)
Would a CNL interface enable someone who is unfamiliar with KR or KR languages to build a knowledge base without help?

This paper presents empirical observations on a Controlled Natural Language (CNL) authoring task for two OWL experts and four OWL novice participants who were learning the CNL known as OWL Simplified English, or OSE \cite{Power2012}. Participants were shown three video tutorials on OSE, each followed by a 10-minute exercise during which they used the SWAT Editing Tool\footnote{http://mcs.open.ac.uk/nlg/SWAT/editor.html}  to construct OSE sentences from domain knowledge in the form of a paragraph of text taken from a Simple Wikipedia article. In a sense, they were performing a translation exercise to convert natural language (English text) into OSE and thus directly into OWL (through typing OSE into the editor interface). 

This task was particularly difficult because the knowledge to be encoded in CNL was not \emph{artificially prepared} by the experimenters but a \emph{naturally-occurring} text written by wikipedia authors. It was thus a radical departure from the kinds of data supplied in other evaluations (see section \ref{relWork}). Our motivation to use such data was that it more closely represents the kind of knowledge that a domain expert might carry in his/her head, i.e., a genuine example of domain data `from the wild'. Consequently, it presented an additional burden on participants because some parts of the source text could \emph{not} be expressed in OWL (or OSE) and some parts were not in a convenient form, therefore participants had to select and organise information as well as encoding it in CNL. 

Knowledge engineering in CNL is a complex task requiring such a large number of skills (or competencies) that it seems unlikely that someone who knows nothing of the underlying formal semantics could be expected to perform well. We break down the requisite skills into three areas (knowledge representation, sentence construction, and identifier name construction).
In observing participants' actions from screen recordings with eye-tracking, our aim was to find out how exactly they modelled the domain knowledge from the text, how they went about constructing ontology axioms and identifier names, and whether they encountered problems whilst doing so.  From our analysis of the screen recordings, we present some insights about their attempts to learn the CNL and construct a knowledge base.  From these, we make predictions about the difficulties that novices, in particular, face and hence the feasibility of CNL as an interface for novices and experts.

\section{Related Studies}
\label{relWork}

Our exploratory study differed radically from other evaluations of CNL knowledge editors in that the material it provided for participants as `knowledge to be encoded' was \emph{not} artificial. We provided a naturally-occurring, human-authored text; other evaluations provided participants with artificial `knowledge', e.g., schematic diagrams \cite{kuhn2009,kuhn2010,kuhn2013},  or  NL sentences contrived with different phrasing and wording from that of the CNL. For example, Funk et al.'s evaluation of the CLOnE language for semantic web ontology editing \cite{Funk2007} gave participants  sentences such as,  `Create a subclass \textit{Journal} of \textit{Periodical}.'  Hallett, Power and Scott \cite{Hallett2007} gave their participants artificial  texts for the task of constructing SQL queries, e.g., `How many patients who received surgical treatment for malignant neoplasm of the central portion of the breast had no curative radiotherapy?'  Garc\'{i}a-Barriocanal et al. \cite{Garcia2006} provided what we assume was an artificially-contrived text for the task of constructing a small ontology. An exception is the study of Laing et al. \cite{Liang2013} which used short texts written by ontology engineers describing a few OWL statements. The major differences between all of these texts and our text is that with the artificial texts and Laing et al.'s texts, participants were provided with convenient terms and, more importantly, only with data that \emph{could be} successfully encoded; whereas, some of our naturally-occurring text \emph{could not be}  encoded, nor were identifier names provided in a convenient form. 

These are important differences because they made more realistic domain experts of our participants, assuming that knowledge inside a domain expert's head is not conveniently organised in a form that would lend itself to CNL encoding.  Thus we forced our participants to \emph{select and reorganise  knowledge} before encoding it. On the other hand, the studies above were focussed on particular competencies (e.g., one aspect of Funk et al.'s study tested whether users had learnt the CNL sentence pattern for expressing a subclass relationship), whereas the purpose of our study was to explore which competencies are important for the task of encoding knowledge in CNL. 

An exception was a study in which participants were encouraged to find encyclopedia articles from which to encode geographical knowledge \cite{Kaljurand}. Because of the constraints of the system used, several hundred domain vocabulary names had to be prepared in advance.  Using the vocabulary provided, participants were able to choose to encode any geographical information they wanted. Unsurprisingly, this produced differing contents that were hard to compare.  However, as in our study, participants used different modelling styles to represent similar information according to their different views of the world and had difficulties producing syntactically acceptable formulations.

Studies exist that compare new ontology editors to popular alternatives like Prot\'{e}g\'{e}, e.g., \cite{HerFerDuc2012}. This was not the aim of our study, which is concerned with the details of learning a CNL for knowledge editing, not with the broader issue of which approach is best.  

\section{Tools, Materials and Method}

\subsection{OWL Simplified English}
OWL Simplified English (OSE), [15], is a relatively free-form language in which each
sentence expresses an OWL statement, and entity names (for individuals, classes and properties) are recognised by their relationship to a handful of common English keywords such as `the', `is', `has', and `a', with minimal classification of content words. It is left to the writer to decide whether to create text that would be recognisable
or understandable as natural English. For instance, in the sentence `A dog is an animal.', text
between `A' and `is' is interpreted as a class name. Likewise, `animal' is a class name
because it is delimited by `aní and `.'. Thus `A because because is an of of of.' would also be a valid OSE sentence, meaning that the class `because because' is a subclass of the class `of of of'.\footnote{A tutorial is available at  mcs.open.ac.uk/nlg/SWAT/EditingToolApril2012/tutorial.pdf}

OSE is relatively unconstrained when constrasted with other CNLs which require
predefined vocabularies. Because the grammar is finite-state,
sentences can quickly be verified as correct, and interpreted in OWL. The language disallows
sentence patterns using connectives like `and', `or', `that', which people would
interpret as structurally ambiguous.

\subsection{SWAT Editing Tool}

\begin{figure}[h]
\begin{center}
\includegraphics[width=4.8in]{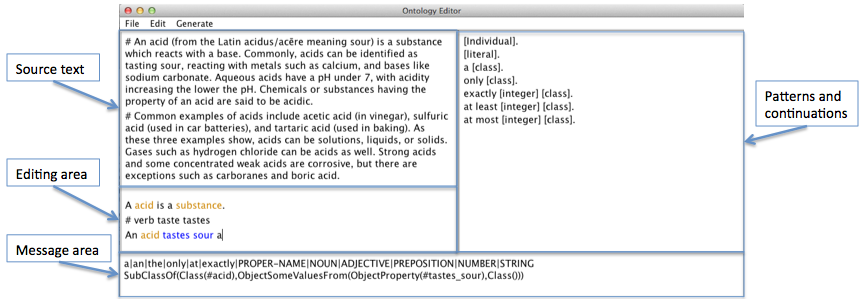} 
\caption{SWAT Editing Tool }
\label{editor}
\end{center}
\end{figure}

The  editing tool\footnote{Downloadable from http://mcs.open.ac.uk/nlg/SWAT/editor.html} used in this study was developed for the SWAT project\footnote{Semantic Web Authoring Tool (SWAT) project funded by the Engineering and Physical Science Research Council (grant no. G033579/1). } as described in Power \cite{Power2013}. It implements OSE  \cite{Power2012}, building OWL statements  dynamically as the user types OSE sentences. Its predictive interface provides sentence patterns (as full or partial sentences) and feedback on the OWL statement being built. As it is typed, text is parsed character-by-character and automatically coloured brown for class names, purple for individuals, blue for properties, and green for literals. Figure \ref{editor} shows a screenshot of the editor set up for the study with the source text inserted as a comment at the top of the editing pane (which was larger than shown here), a context-sensitive list of allowed sentence or continuation patterns (RHS), and a context-sensitive message area for dynamic feedback.   

In the editing area, a sentence, `A acid is a substance.', containing an English grammatical error has been, nonetheless, accepted by the OSE finite-state parser. Another sentence is being typed, `An acid tastes sour a \dots', the message area shows the partially-constructed OWL statement in which `tastes sour' is recognised as a property name (possibly the author intended `sour' as a literal, denoted by double quotes in OSE). 

\subsection{Materials}
\noindent \textbf{Instructional videos}
To ensure that all participants received identical tuition, we recorded three 5- to 8-minute instructional videos with screen recordings and spoken commentaries demonstrating how to construct OSE sentences and identifier names\footnote{Videos may be viewed at http://mcs.open.ac.uk/nlg/SWAT/editor.html}.

Additionally, participants were given `crib' sheets summarising all sentence patterns taught in the videos. In video 1, participants were taught to construct three types of OSE sentences (class subsumption, class membership, and disjoint classes) using class (concept) names and individual (class member) names. In video 2, participants were taught to construct multi-word property and individual names, literals in quotes for previously-taught sentences, and new sentence patterns for existential property restrictions. In video 3, participants were taught to construct sentence patterns for equivalent classes, property restrictions  `only', `exactly, `at least', `at most', and property domains and ranges.

\noindent \textbf{Text to be `translated'}
The source text containing the domain knowledge was from a Simple Wikipedia\footnote{Downloaded from http://simple.wikipedia.org/wiki/Acid on 29th November 2012} article about acid:
\begin{quotation}
\noindent{\scriptsize An acid (from the Latin acidus/ac\={e}re meaning sour) is a substance which reacts with a base. Commonly, acids can be identified as tasting sour, reacting with metals such as calcium, and bases like sodium carbonate. Aqueous acids have a pH under 7, with acidity increasing the lower the pH. Chemicals or substances having the property of an acid are said to be acidic.}

{\scriptsize Common examples of acids include acetic acid (in vinegar), sulfuric acid (used in car batteries), and tartaric acid (used in baking). As these three examples show, acids can be solutions, liquids, or solids. Gases such as hydrogen chloride can be acids as well. Strong acids and some concentrated weak acids are corrosive, but there are exceptions such as carboranes and boric acid.}
\end{quotation}

We chose this as an appropriate expository text because it presented the typical difficulties
and ambiguities of naturally-occurring text while avoiding content requiring past tense (e.g., events in history). To check the suitability of the source text, the authors independently tried to recast its content in OSE, and produced three texts showing somewhat different modelling styles (see also section \ref{genDisc}). 

\noindent \textbf{Task instructions}
Participants were given written instructions as follows:
\begin{quotation}
\noindent{\scriptsize Your task is to enter information about classes and individuals from the text that you have been given using the sentence patterns shown in the tutorial. Try to use only information that you find in the text but you may use alternative phrases if you wish. There will be information in the text that you do not know how to express yet; do not worry, just leave it out for now. You will be adding more after the next tutorial.}
\end{quotation}

\subsection{Method}

Six participants, two OWL experts and four novices, completed the study individually in a small room in the Open University human interaction laboratory supervised by the first author who calibrated the eye-tracker, started the videos, controlled the timings of each 10-minute CNL authoring session, and saved participants' OSE text files. 

\section{Results}

OSE texts written and edited by participants range from 3 to 30 sentences.\footnote{Available from mcs.open.ac.uk/nlg/SWAT/WilliamsEtAl2014-ParticipantTexts.pdf}  Overall, it is surprising how much text they managed to write. Even though the resulting knowledge modelling in OWL is sometimes not ideal,\footnote{We chose \emph{not} to view any particular modelling style as being `correct'.} nevertheless, all participants except one managed to construct a simple ontology about acids.

\subsection{Learning OSE}
Analysis of screen recordings of participants' authoring sessions revealed that all quickly learnt the patterns `A [class] is a [class].' and  `[Individual] is a [class].' Experts seemed to pick up the controlled language with remarkable ease.

Some novices attempted to write in natural English rather than OSE, e.g., `A is B, so is C.' and a conjunction in the subject NP `A and B are \dots'. A novice had problems with verb tenses (s/he tried to use the past tense form) and with plural nouns when the singular is required. Further problems are noted in the following subsections.

\subsection{Knowledge Modelling}

\noindent\textbf{Building class hierarchies:} A major difference between experts and novices was that \emph{experts had one more level of depth in their hierarchies}. Experts identified more subclasses than novices who would typically fail to state, e.g., that strong, and concentrated weak acids are types of acid. An expert would include these and also `base' and `metal' as subclasses of `substance', and `aqueous acid' as a subclass of `acid'.
 
\noindent\textbf{Defining class members:} Regarding class membership, everyone constructed `Acetic acid is an acid', `Tartaric acid is an acid', and so on from the original sentence where these are clearly stated: `Common examples of acids include acetic acid (in vinegar), sulfuric acid (used in car batteries), and tartaric acid (used in baking).'

\noindent\textbf{Explicit vs. implicit information:} In general, where class subordination and class membership information is \emph{explicit} in the source text, all participants managed to model it; however, where information is \emph{implicit}, only experts modelled it.  

\noindent\textbf{Constructing relationships (properties) between individuals:} All participants managed to construct at least one property; however, nearly everyone had problems constructing property names (see section \ref{resultsIds}). Some novices had problems attempting to construct relationships in the text that are difficult, or impossible, to model in OWL, e.g., the vague `can be' in `Gases such as hydrogen chloride can be acids'. 

\noindent \textbf{Translating source text information:} There is evidence that everyone tried to translate directly from the source text. All participants started constructing their ontologies with some variation of the sentence `An acid is a substance.' This corresponds with part of the first sentence in the original text, \textit{`An acid (from the Latin acidus/ac\={e}re meaning sour) is a substance which reacts with a base.'} One person even copied the sentence from the original text, pasted it into the editor and deleted redundant parts of it. Often, they wrote many OSE sentences for a single source text sentence; indeed, expert E7 wrote six for the first source text sentence, perhaps exploring the range within OSE.

\begin{table}[h]
\footnotesize 
\begin{center}
\tabcolsep=0.11cm
\begin{tabular}{|l|c|c|c|c|c|c|}
\hline 
\textbf{OSE Sentence} & \textbf{N4} & \textbf{N5} & \textbf{N6} & \textbf{N2} & \textbf{E3} & \textbf{E7} \\
\hline 
Aqueous acids are under 7. & &  & & \checkmark &  & \\
Aqueous acids has ph under 7. & &  &\checkmark &  &  & \\
An aqueous acid is defined as a liquid that has pH below 7.& &  & &  &  & \checkmark \\
An acid has as pH ``7 or less". &  &  &  &  & \checkmark  &  \\
Acid has ph under 4. & &  &\checkmark &  &  & \\
Acid is definied as a substance  that has ph under  7. &  &  & \checkmark &  &  &  \\
An acidic is defined as a subtance  that has as property of an acid. &  &  &  &  & \checkmark &  \\
Acidity is inverted proportion  to a ph.&  &  & \checkmark &  &  &  \\
 \hline 
\end{tabular}
\end{center}
\caption{\label{translating3} Modelling the sentence \textit{`Aqueous acids have a pH under 7, with acidity increasing the lower the pH. Chemicals or substances having the property of an acid are said to be acidic.'} }
\end{table}

Participants demonstrated surprising consensus in modelling the second source text sentence (apart from class-individual differences). The screen recordings revealed that most struggled to interpret the vague term `commonly' in `commonly, acids can be identified as tasting sour', encoding the meaning as `\emph{all} acids taste sour'. 

Again, screen recordings revealed difficulties with constructing a property for `pH' (all participants that attempted it had difficulties). See also table \ref{translating3} for a comparison of their attempts.

In modelling usages of acids and whether they are solids, liquids or gases, only novice N2 and expert E7 attempted to model usages of common acids but N2 seemingly misunderstood the universal restriction by writing `Tartaric acid is used only in baking'. Expert E3 provided a plausible disjoint union for solution, liquid, solid, and gas classes.

As a final example, consider how participants modelled the knowledge that acids can be weak or strong, or corrosive or non-corrosive. E7 was inventive in writing `Boric acid corrodes exactly 0 substances.'  N2 specified `non-corrosive' and `corrosive' classes with the latter equivalent to `acid'. If, as indicated by their names, s/he had made `corrosive' and `non-corrosive' disjoint, a non-corrosive acid could not logically exist. 

\subsection{Sentence Construction}

\noindent\textbf{Sentence pattern usage} Table \ref{patterns} shows a breakdown of sentence pattern usage by participant. It is clear that the OWL experts, E3 and E7, produced more sentences than OWL novices (N4, N5, N6 and N2). The mean number is 26 for experts and 15 for novices not including N4.  \emph{Experts attempted a greater variety and more complex patterns than novices}. Eleven of the patterns taught in the videos were used, six were taught but not used, and a further seven patterns were used that were not taught.

\begin{table*}[ht]
\scriptsize 
\center
\tabcolsep=0.11cm
\begin{tabular}{|l|r|r|r|r|r|r|r|}
\hline 
\textbf{Pattern}  & \textbf{N4} & \textbf{N5} & \textbf{N6} & \textbf{N2} & \textbf{E3} & \textbf{E7} & \textbf{Total} \\
\hline 
{[}Individual] is a [class].   & 1 & 7 & 6 & 10 & 7 & 9 & 40 \\
A [class] is a [class].   & 1 & 2 & 1 & 3 & 6 & 7 & 20 \\
No [class] is a [class].   &  & 3 &  &  & 3  & 1 & 7 \\
A [class] [has-property] a [class].   &  &  &  & 1 & 3 & 2 & 6 \\
{[}Individual] [has-property] [Individual].  & 1 &  & 1 &  &  & 2 & 4 \\
{[}Individual] [has-property]  a [class].  &  &  & 3 &  &  & 1 & 4 \\
{[}Individual] [has-data-property] [literal].  &  &  & 2 & 1 &  &  & 3 \\
A [class] [has-property] [Individual].  &  &  &  &  &  & 3 & 3 \\
{*[}Individual] is defined as a [class]. &  &  &  & 3 &  &  & 3 \\
A [class] is defined as a [class] that [has-property] a [class].   &  & 1 &  &  & 1 & 1 & 3 \\
{[}Individual] [has-property]  a [class] that [has-data-property] [literal].  &  &  & 1 &  &  &  & 1 \\
{[}Individual] [has-property] only [class].  &  &  &  & 1 &  &  & 1 \\
{[}Individual] [has-property] exactly [integer] [class].  &  &  &  &  &  & 1 & 1 \\
A [class] is defined as a [class] that [has-data-property] [literal]. &  &  &  &  &  & 1 & 1 \\
A [class] is a [class] or a [class] or a [class] or a [class].  &  &  &  &  & 1 &  & 1 \\
 A [class] [has-data-property] [literal].  &  &  &  &  & 1 &  & 1 \\
A [class] [has-property] at least [integer] [class].  &  &  &  &  &  & 1 & 1 \\
Anything that [has-property] something is a [class].  &  &  &  &  &  & 1 & 1 \\
\hline 
\textbf{Total Sentences} & \textbf{3}  & \textbf{13} & \textbf{14} & \textbf{19} & \textbf{22} & \textbf{30} & \textbf{101} \\
\hline 
\textbf{Total Unique patterns} & \textbf{3} & \textbf{4} & \textbf{6} & \textbf{6} & \textbf{7} & \textbf{12} & \textbf{18} \\
\hline
\end{tabular}
\footnotesize
\caption{\label{patterns}  Sentence pattern frequencies in participants' final texts (*error in editor)}
\end{table*}

\noindent\textbf{Understanding that OSE sentences must conform to syntactic rules and ability to correct errors:} There is evidence in the screen recordings that participants noticed when sentences were incorrect and tried to correct them. Eye tracks and gaze duration circles over sentences being written before and after adding a full stop seem to indicate sentence checking. Sometimes a sentence pattern was selected in what looked like an attempt to correct a half-written sentence. On the other hand, there was often little attempt to conform to English grammar rules (`a acid' was not corrected to `an acid' by three participants); indeed OSE does not recognise (or colour-code) them as errors.

Some participants failed to correct sentences with syntactic flaws that could not be parsed by the finite state automaton in the editor.  `An acid tastes Sour and reacts with metals.' was produced by N4 who had successfully declared `tastes' and `reacts' as verbs but failed to remember that if it were a class, `metal' should be singular with a determiner, or if an individual, it should be capitalised. Consequently, the entire phrase `reacts with metals' was treated as a property name by the editor so the sentence was incomplete. Similar errors were produced by other novices.

\noindent\textbf{Eye-tracking during sentence construction} Table \ref{durations} shows the proportion of total visit duration times, i.e., total times that the eye tracker recorded the participant looking at the source text, editing area, patterns and continuations area and message area (see figure \ref{editor}). Data for E7 are missing; the eye tracker did not work for this participant. The table shows times for the first exercise only, because most participants had written enough material by the second exercise to start scrolling the editing pane and thus the source text and editing area were no longer fixed inside the relevant areas marked for automatic calculation of visit duration. From the videos, we observed that \emph{most participants spent a long time re-reading text that they had already written} (especially E3), perhaps checking consistency or for missing information. This observation only accounts for part of the total visit durations, however, since time was also spent composing and editing text. Some people looked at messages and OWL statements in the message pane, but these did not receive much attention overall (zero or 1\% of total visit duration).

\begin{table}[h]
\footnotesize
\center
\tabcolsep=0.11cm
\begin{tabular}{|l|r|r|r|r|r|}
\hline 
\textbf{Part of Editor} & \textbf{N4} & \textbf{N5} & \textbf{N6} & \textbf{N2} & \textbf{E3}  \\
\hline 
Source text & 17\% & 25\% & 54\% & 45\% & 27\%   \\
Editing area  & 54\% & 47\% & 30\% & 48\% & 72\%   \\
Sentence Patterns & 28\% & 28\% & 16\% & 6\% & 1\%   \\
Message Pane & 1\% & 1\% & 0\% & 1\% & 1\%   \\
\hline 
\end{tabular}
\footnotesize
\caption{\label{durations} Total visit durations (1st exercise only)}
\end{table}

\subsection{Identifier Name Construction}
\label{resultsIds}

Table \ref{names} shows a breakdown of identifier names in the final CNL texts by type (class names, individual names, or property names) and participant.  All participants successfully produced at least one of the three types, and all produced 19 to 25 different names except N4. There was considerable variety in the identifiers constructed 
with 58 unique names amongst 119 total, where type is treated as a difference (e.g., `acid' the class name is counted as different from `acid' the named individual).  

\noindent \textbf{Modifications to names from the source text} 
Table \ref{names} shows frequencies of class, individual and property names and a breakdown whether these are identical to, or modified from, the source text, or were not present in the source text. It is immediately apparent that most identifier names were derived from words and phrases in the source texts (94\%). Surprisingly, although most identifiers were similar to terms in the source text, only around half had exactly the same morphological forms.  Creation of entirely new terms, such as synonyms of source text terms (`below' from `under') or antonyms ('non-corrosive' from `corrosive') was rare, only  6\% used other English phrases; this could be because participants thought that using alternative words would change the meaning, or because it requires greater mental effort.

With \emph{class names}, almost all modifications to source text terms consisted of changing plural nouns into singular nouns (32 of 33 modifications, or 97\%), e.g., `gases' to `gas' and `car batteries' to `car battery'.  This evidence indicates that plural-to-singular noun modification presented no difficulties. Other modifications were construction of a new term, `non-corrosive', not in the source, and conversion of the progressive verb `tasting' into the noun `taste'.  Fewer than half, or 24 of the total 59 class names, (41\%) were identical to strings in the source text.

Conversely, \emph{named individual identifiers} are almost all identical to strings in the source text (35 out of 39, 90\%); this was expected since most were names of chemical compounds. Of those that were not, one was a plural noun made singular, two were separated adjectives and nouns, and the other, `s-acid', did not exist in the source text.

As for \emph{property names}, all except one were different from strings in the source text; 16 out of 21, 76\%, were different. The majority of modifications were the insertion of `is' or `has' before a noun and optional preposition (83\% of modifications), e.g., `is used in', `has common taste' (14, including those containing nouns that were not in the source text). This type of name is commonly used by ontology authors (Power, 2010; Power and Third, 2012, Williams, 2013) and, indeed, it was taught in our tutorials. Other modifications were varied, including changing the progressive verb `tasting' to the noun `taste' or to the verb `tastes' and the adjective `corrosive' to the 3rd person present singular verb `corrodes', and the verb `reacts' to the noun `reactant'. 

\begin{table}[h]
\footnotesize
\center
\tabcolsep=0.11cm
\begin{tabular}{|l|c|c|c|c|c|c|r|r|r|r|}
\hline 
\textbf{Type} & \textbf{N4} & \textbf{N5} & \textbf{N6} & \textbf{N2} & \textbf{E3} & \textbf{E7} & \textbf{Total} & \textbf{Identical to}  & \textbf{Modified}  & \textbf{Other} \\
&&&&&&&&\textbf{source text}  & \textbf{source text}  & \\
\hline 
Class  & 2 & 13 & 6 & 13 & 15 & 10 & 59 & 24 (41\%) & 33 (56\%) &  2 (3\%)\\
Indiv & 2 & 7 & 9 & 7 & 6 & 8 & 39 & 35 (90\%) & 3 (8\%) &  1 (2\%)\\
Prop & 1 & 2 & 4 & 3 & 4 & 7 & 21 & 5 (24\%) & 12 (57\%)  & 4 (19\%) \\
\hline 
\textbf{Total} & \textbf{5} & \textbf{22} & \textbf{19} & \textbf{23} & \textbf{25} & \textbf{25} & \textbf{119} & 64 (54\%) & 48 (40\%)   & 7 (6\%) \\
\hline
\end{tabular}
\footnotesize
\caption{\label{names} Frequencies and origins of identifier names by type.}
\end{table}

\noindent \textbf{Difficulties}
Three OWL novices had difficulty understanding the difference between classes and individuals.
They constructed `acid' as both a class and an identifier (even though the editor colours them differently and shows their different OWL expressions).

Constructing multi-word names and understanding how quotes are used was another difficulty. OSE uses quotes for two different purposes: (i) class names containing keywords such as `and', and (ii) literals. Some participants had initial difficulties, however, everyone except N4 managed to create multi-word names, e.g., Boric acid, Hydrogen chloride.  N4 seemed to have the idea that multi-word names could not be used, hence his/her attempts to use quotes and camel case `reactsWith'.

A third difficulty was constructing property names; evidence from failed attempts in the screen recordings showed that all participants experienced some difficulty. All except N4 managed to use the OSE syntax for declaring a verb, e.g., `\#verb react reacts'.

\noindent\textbf{Participants' comments}
Regarding the task itself, participants did not mention their difficulties constructing syntactically correct sentences. One commented that it is hard to build an ontology without a particular application in mind. Regarding OSE, some participants were interested in how exactly each sentence is parsed. A participant who is a computer programmer noted that the syntax of OSE seemed more complex than a programming language. Other comments tended to be about the user interface: it should give more help; provide better handling of placeholders in generic sentence patterns; and display new verbs immediately in the options list. One expert Prot\'{e}g\'{e} user requested a display of the complete OWL ontology and class hierarchy rather than just the statement under construction.

\section{General Discussion}
\label{genDisc}

 \noindent \textbf{Is a CNL easy to learn?}
Nearly everyone in the study was able to quickly learn to: construct simple sentence patterns; make use of words and phrases (suitably modified) from the source text to create identifier names; declare verbs for use in property names; and correct at least some syntactic errors. All participants made strikingly similar modifications to source text phrases, converting plural nouns to singular for class names and inserting `is' or `has' before nouns to form property names. 

Compared with novices, OWL experts produced a larger number of well-formed OSE sentences, utilising a wider range of patterns. We assume that this was not because they were better at learning the syntax of the language, but because they were more familiar with the KR task.

All participants had difficulty constructing sentences with properties.  Some novices, in particular, tended to \emph{avoid} properties by introducing more classes, e.g., rather than `An acid corrodes a metal.', they would write `An acid is a corrosive substance.'  

Participants spent a lot of time reading previously-written text. Perhaps they were looking to see what worked before, in the same way that programmers search for code examples. If so, it suggests that providing many examples of well-formed sentences might benefit OSE learners.

Regarding differences in modelling, we are aware that styles differ; it is unclear whether, for instance, `tartaric acid' should be modelled as an instance or as a subclass of `acid'. We therefore decided \emph{not} to treat any particular model as `correct'.    Likewise, we choose a naturally occurring source text expecting that it would elicit different models (since we tried the exercise ourselves before the experiment). In an ideal world, domain experts would collaborate with knowledge engineers to build ontologies; CNLs such as OSE could provide a useful communication medium between the two.

Although our focus was on learning OSE, user interface issues emerged, particularly lack of attention to the message pane suggesting that it should be re-positioned.

\noindent \textbf{Would a CNL enable someone unfamiliar with KR to build a KB?}
All participants largely agreed on class subsumption and membership \emph{explicitly} present in the source text, demonstrating that certain aspects of building a KB are accessible to everyone. However, a marked difference between OWL experts and novices was the greater organisation and depth of experts' class hierarchies. Experience with knowledge engineering enabled the experts to model knowledge that was not \emph{explicit}  in the source text but \emph{implied}. OWL novices did not model implict knowledge, perhaps indicating that they did not realise that the implied subclass relationship between, say, strong acid and acid, so obvious to a human reader, must be specified.  See also Third  \cite{Third2012}.

OWL novice errors noted by Rector et al. \cite{rector2004} were: (i) failure to make `hidden information' in identifier names explicit, (ii) misunderstanding the universal restriction, (iii) misuse of logical `and' and `or', (iv) ignorance of the open-world assumption (and consequent failure to specify disjoint classes), and (v) incorrect placement of logical `not'. \emph{OWL novices} in our study made errors (i) and (iv). Features in (ii), (iii),  and (v) were little used. 

Novices in our study made the error of modelling the same thing as a class \emph{and} an individual; therefore, to Rector et al.'s list we would add (vi) confusion of \emph{general} concepts (classes) with \emph{specific} instances of the classes (`individuals' in OWL). 

\section{Conclusion and Future Work}

While OWL experts seemed to master OSE quickly and produced small ontologies with ease. Clearly, novices experienced difficulties and require more guidance such as examples of syntactically correct sentences. 

Alternative interfaces such as \textsc{WYSIWYM} \cite{Power1998} have achieved some success with novices at the expense of freedom to type text as in a conventional editor. Dialogue systems currently under development, e.g., in the WhatIf! project \cite{parvizi2013}, might provide a way forward.  If a system were to have the ability to respond with intelligent and appropriate questions and remarks about possibly unintended entailments present in knowledge entered, it might enable even novices to gain some insight into the formal semantics and hence construct KBs that are logically consistent.

\bibliographystyle{splncs03}

\bibliography{CNL-final}

\end{document}